\documentclass[runningheads,a4paper]{llncs}

\usepackage{amssymb}
\usepackage{graphicx}

\usepackage{color}

\usepackage{url}
\urldef{\mailsa}\path|{alfred.hofmann, ursula.barth, ingrid.haas, frank.holzwarth,|
\urldef{\mailsb}\path|anna.kramer, leonie.kunz, christine.reiss, nicole.sator,|
\urldef{\mailsc}\path|erika.siebert-cole, peter.strasser, lncs}@springer.com|    
\newcommand{\keywords}[1]{\par\addvspace\baselineskip
\noindent\keywordname\enspace\ignorespaces#1}

\begin{document}

\mainmatter  

\title{Automatic Parameter Adaptation for Multi-Object Tracking}

\titlerunning{Automatic Parameter Adaptation for Multi-Object Tracking}

%
%

\author{Duc Phu CHAU \and Monique THONNAT \and Fran\c cois BREMOND \\
\email{\{Duc-Phu.Chau, Monique.Thonnat, Francois.Bremond\}@inria.fr}
}

\authorrunning{D. P. Chau et al.}

\institute{STARS team, INRIA Sophia Antipolis, France \\
\texttt{http://team.inria.fr/stars}
}

%
%

\toctitle{Automatic Parameter Adaptation for Multi-Object Tracking}
\tocauthor{Duc Phu CHAU, Monique THONNAT, and Fran\c cois BREMOND}
\maketitle

\begin{abstract}

Object tracking quality usually depends on video context (e.g. object occlusion level, object density). In order to decrease this dependency, this paper presents a learning approach to adapt the tracker parameters to the context variations. In an offline phase, satisfactory tracking parameters are learned for video context clusters. In the online control phase, once a context change is detected, the tracking parameters are tuned using the learned values. The experimental results show that the proposed approach outperforms the recent trackers in state of the art. This paper brings two contributions: (1) a classification method of video sequences to learn offline tracking parameters, (2) a new method to tune online tracking parameters using tracking context.

\end{abstract}

\section{Introduction}
\label{secIntro}

Many approaches have been proposed to track mobile objects in a scene. However the quality of tracking algorithms always depends on scene properties such as: mobile object density, contrast intensity, scene depth and object size. The selection of a tracking algorithm for an unknown scene becomes a hard task. Even when the tracker has already been determined, it is difficult to tune online its parameters to get the best performance.

Some approaches have been proposed to address these issues. The authors in \cite{kuo} propose an online learning scheme based on Adaboost to compute a discriminative appearance model for each mobile object. However the online Adaboost process is time consuming. In \cite{borji12}, the authors present an online learning approach to adapt the object descriptors to the current background. However, the training phase requires the user interaction. This increases significantly the processing time and is not practical for the real time applications.

Some approaches integrate different trackers and then select the convenient tracker depending on video content \cite{prost}\cite{yoon12}. These approaches run the tracking algorithms in parallel. At each frame, the best tracker is selected to compute the object trajectories. These two approaches require the execution of different trackers in parallel which is expensive in terms of processing time. In \cite{dpchauIcdp11}, the authors propose a tracking algorithm whose parameters can be learned offline for each tracking context. However the authors suppose that the context within a video sequence is fixed over time. Moreover, the tracking context is manually selected.

These studies have obtained relevant results but show strong limitations. To solve these problems, we propose in this paper a new method to tune online the parameters of tracking algorithms using an offline learning process. In the online phase, the parameter tuning relies entirely on the learned database, this helps to avoid slowing down the processing time of the tracking task. The variation of context over time during the online phase is also addressed in this paper.

This paper is organized as follows. Section 2 and 3 present in detail the proposed approach. Section 4 shows the results of the experimentation and validation. A conclusion as well as future work are given in the last section.

\section{Offline Learning}
\label{secOfflineLearning}

\begin{figure}[b]
\center
\includegraphics[width=12cm]{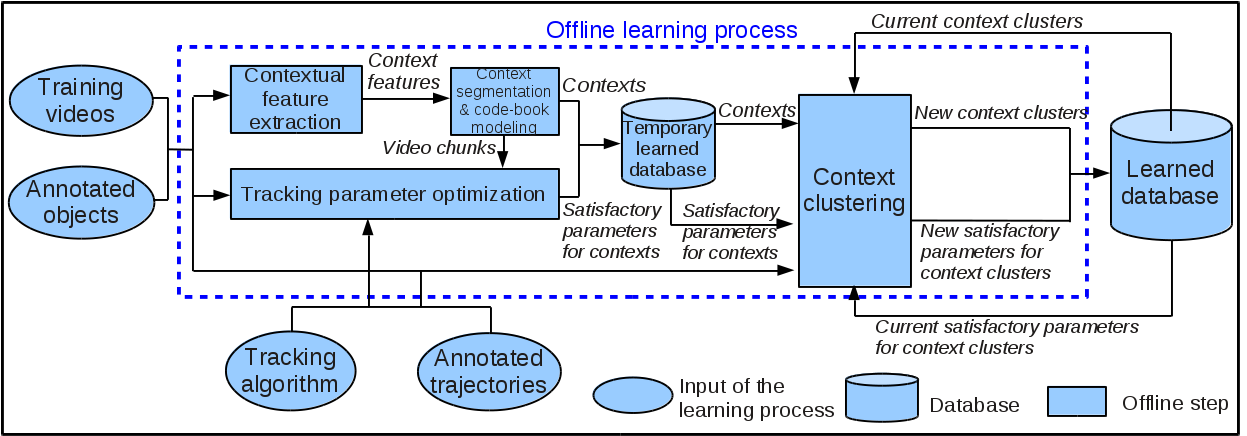}
\caption{The offline learning scheme}
\label{fig_learning_scheme}
\end{figure}

The objective of the learning phase is to create a database which supports the control process of a tracking algorithm. This database contains satisfactory parameter values of the controlled tracking algorithm for various scene conditions. This phase takes as input training videos, annotated objects, annotated trajectories, a tracking algorithm including its control parameters. The term ``control parameters'' refers to parameters which are considered in the control process (i.e. to look for satisfactory values in the learning phase and to be tuned in the online phase). At the end of the learning phase, a learned database is created. A learning session can process many video sequences. Figure \ref{fig_learning_scheme} presents the proposed scheme for building the learned database.

The notion of ``context'' (or ``tracking context'') in this work represents elements in the videos which influence the tracking quality. More precisely, a context of a video sequence is defined as a set of six features: density of mobile objects, their occlusion level, their contrast with regard to the surrounding background, their contrast variance, their 2D area and their 2D area variance. For each training video, we extract these contextual features from annotated objects and then use them to segment the training video in a set of consecutive chunks. Each video chunk has a stable context. The context of a video chunk is represented by a set of six code-books (corresponding to six features). An optimization process is performed to determine satisfactory tracking parameter values for the video chunks. These parameter values and the set of code-books are inserted into a temporary learned database. After processing all training videos, we cluster these contexts and then 
compute satisfactory tracking parameter values for context clusters.

In the following, we describe the four steps of the offline learning process: (1) contextual feature extraction, (2) context segmentation and code-book modeling, (3) tracking parameter optimization and (4) context clustering.

\subsection{Contextual Feature Extraction}

For each training video, the context feature values are computed for every frame.

\textbf{1. Density of Mobile Objects:} A high density of objects may lead to a decrease of object detection and tracking performance. The density of mobile objects at $t$ is defined by the sum of all object areas over the 2D camera view.

\textbf{2. Occlusion Level of Mobile Objects:} An occlusion occurrence makes the object appearance partially or completely invisible. The occlusion level of mobile objects at instant $t$ is defined as the ratio between the 2D overlap area of objects and the object 2D areas.

\textbf{3. Contrast of Mobile Objects:} The contrast of an object is defined as the color intensity difference between this object and its surrounding background. An object with low contrast decreases the discrimination of the appearance between different objects. The contrast of mobile objects at instant $t$ is defined as the mean value of the contrasts of objects at instant $t$. 

\textbf{4. Contrast Variance of Mobile Objects:} When different object contrast levels exist in the scene, a mean value cannot represent correctly the contrast of all objects in the scene. Therefore we define the variance of object contrasts at instant $t$ as their standard deviation value.

\textbf{5. 2D Area of Mobile Objects:} 2D area of an object is defined as the number of pixels within its 2D bounding box. Therefore, this feature characterizes the reliability of the object appearance for the tracking process. The 2D area feature value at $t$ is defined as the mean value of the 2D areas of mobile objects at instant $t$.

\textbf{6. 2D Area Variance of Mobile Objects:} Similar to the contrast feature, we define the variance of object 2D areas at $t$ as their standard deviation value.

\subsection{Context Segmentation and Code-book Modeling}
\label{sec_context_segmentation}

The contextual variation of a video sequence influences significantly the tracking quality. Therefore it is not optimal to keep the same parameter values for a long video. In order to solve this issue, we propose an algorithm to segment a training video in consecutive chunks, each chunk is defined as having a stable context (i.e. the values of a same context feature in each chunk are close to each other). This algorithm is described as follows.

First, the training video is segmented in parts of $l$ frames. Second, the contextual feature values of the first part is represented by a context code-book model. From the second video part, we compute the distance between the context of the current part and the context code-book model of the previous part. If their distance is lower than a threshold $Th_1$ (e.g. $0.5$), the context code-book model is updated with the current video part. Otherwise, a new context code-book model is created to represent the context of the current video part. At the end of the context segmentation algorithm, the training video is divided into a set of chunks (of different temporal lengths) corresponding to the obtained context code-book models.  The following sections present how to represent a video context with a code-book model; and how to compute the distance between a context code-book model and a context.
\newline

\textbf{1. Code-book Modeling:}
\label{sec_cb_modeling}
During the tracking process, low frequent contextual feature values play an important role for tuning tracking parameters. For example, when mobile object density is high in few frames, the tracking quality can decrease significantly. Therefore, we decide to use a code-book model \cite{codebook} to represent the values of contextual features because this model can estimate complex and low-frequency distributions. In our approach, each contextual feature is represented by a code-book, called \textbf{feature code-book}, denoted $cb^k, \ k\ =\ 1..6$. So a video context is represented by a set of six feature code-books, called \textbf{context code-book model}, denoted $CB$, $CB \ = \{cb^k,\ k = 1..6 \} $. A feature code-book is composed of a set of code-words describing the values of this feature. The number of code-words depends on the diversity of feature values.

\textbf{ Code-word definition: }
\label{sec_def_cw}
A code-word represents the values and their frequencies of a contextual feature. A code-word $i$ of code-book $k$ ($k$ $=$ $1..6$), denoted $cw_i^k$, is defined as follows:
\begin{equation}
\label{eq_define_cw}
 cw_i^k = \{ \overline{\mu_i^k},\ m_i^k ,\ M_i^k,\ f_i^k \}
\end{equation}

\noindent where $\overline{\mu_i^k}$ is the mean of the feature values belonging to this code-word; $m_i^k$ and $M_i^k$ are the minimal and maximal feature values belonging to this word; $f_i^k$ is the number of frames when the feature values belong to this word. For each frame $t$, the code-book $cb^k(k = 1.. 6)$ is updated with the value of context feature $k$ computed at $t$.
\newline

\textbf{2. Context Distance:}
\label{sec_context_distance}
Table \ref{tabFunc} presents the algorithm to compute the distance between a context $c$ and a context code-book model $CB \ = \{cb^k,\ k = 1..6 \} $. The function $distance(\mu^k_t,\ cw_i^k)$ is defined as a ratio between $\mu_t^k$ and $\overline{\mu_i^k}$. This distance is normalized in the interval $[0,\ 1]$. 

\begin{table}[t]
\begin{center}
\begin{small}
\begin{tabular}{|p{12cm}|}
\hline
function $contextDistance(c,\ CB ,\ l)$ \\

Input: context code-book model $CB$, context $c$, $l$ (number of frames of context $c$)\\

Output: context distance between code-book model $CB$ and context $c$	\\ \\

$countTotal = 0$; \\
For each code-book $cb^k$ in $CB$ ($k\ = 1..6$)\\
\hspace{20 pt} $count = 0$;\\
\hspace{20 pt} For each value $\mu_t^k$ of context $c$ at time $t$	\\
\hspace{40 pt}    For each codeword $cw^k_i$ in code-book $cb^k$	\\
\hspace{60 pt}        if ($distance(\mu_t^k,\ cw_i^k)$ $<$  $\epsilon$) \{ $count$++; break; \} \\
\hspace{20 pt} if $(count$ / $l$ $<$ $0.5)$ return $1$;	\\
\hspace{20 pt} $countTotal\ += count;$	\\

return $( \ 1$ $-$ $countTotal/(l*6) \ )$ \\
\hline
\end{tabular}
\end{small}
\end{center}
\caption{\label{tabFunc} Algorithm for computing the distance between a context code-book $CB$ and a video context $c$}
\end{table}

\subsection{Tracking Parameter Optimization}
\label{sec_opt}
The objective of the tracking parameter optimization task is to find the values of the control parameters which ensure the tracking quality higher a given threshold for each video chunk. These parameters are called ``satisfactory parameters''. This task takes as input annotated objects, annotated trajectories, a tracking algorithm, a video chunk and control parameters for the considered tracker. The annotated objects are used as object detection results. Depending on the search space size and the nature of the control parameters, we can select suitable optimization algorithm (e.g. enumerative search, genetic algorithm).

\subsection{Context Clustering}
\label{sec_offline_clustering}

The context clustering step is done at the end of each learning session when the temporary learned database contains the processing results of all training videos. In some cases, two similar contexts can have different satisfactory parameter values because optimization algorithm only finds local optimal solutions. A context clustering is thus necessary to group similar contexts and to compute satisfactory parameter values for the context clusters.

In this work, we decide to use the Quality Threshold Clustering algorithm for this step because this algorithm does not require the number of clusters as input. Once contexts are clustered, all the code-words of these contexts become the code-words of the created cluster. The tracking parameters for a cluster is defined as a combination of tracking parameters belonging to clustered contexts.

\section{Online Parameter Adaptation}
\label{secOnlineParaAdapt}

\begin{figure}[t]
\centering
\includegraphics[width=12cm]{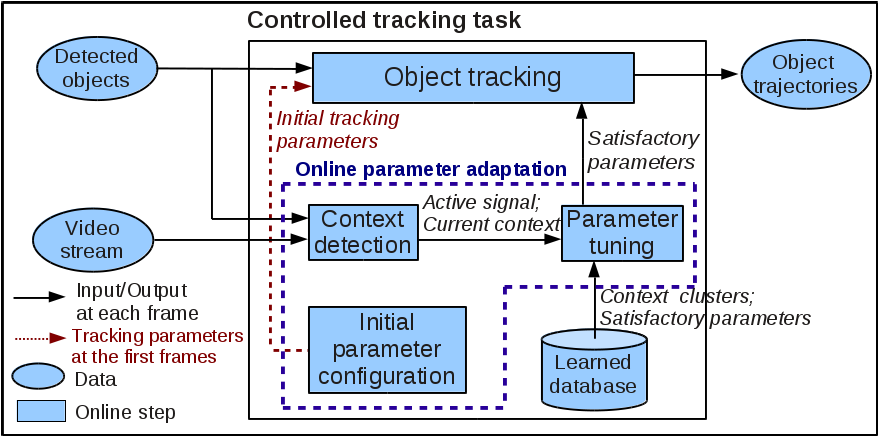}
\caption{The online parameter adaptation scheme}
\label{fig_online_scheme}
\end{figure}

In this section, we describe the proposed controller which aims at tuning online the tracking parameter values for obtaining satisfactory tracking performance. The online parameter adaptation phase takes as input the video stream, the list of detected objects at every frame, the learned database and gives as output the satisfactory tracking parameter values for every new context detected in the video stream (see figure \ref{fig_online_scheme}). In the following sections, we describe the two main steps of this phase: the context detection and parameter tuning steps.

\subsection{Context Detection}
\label{secContextChange}

This step takes as input for every frame, the list of the current detected objects and the image. For each video chunk of $l$ frames, we compute the values of the contextual features. A contextual change is detected when the context of the current video chunk does not belong to the context cluster (clusters are learned in the offline phase) of the previous video chunk. In order to ensure the coherence between the learning phase and the testing phase, we use the same distance defined in the learning phase (section \ref{sec_context_distance}) to perform this classification. If this distance is lower than threshold $Th_1$, this context is considered as belonging to the context cluster. Otherwise, the ``Parameter tuning'' task is activated.

\subsection{Parameter Tuning}
\label{sec_para_adaptation}

The parameter tuning task takes as input an active signal and the current context from the ``context detection'' task, and gives as output satisfactory tracking parameter values. When this process receives an activate signal, it looks for the cluster in the learned database to which the current context belongs. Let $\mathfrak{D}$ represent the learned database, a context $c$ of a video chunk of $l$ frames belongs to a cluster $C_i$ if both conditions are satisfied:
\begin{eqnarray}
\label{eq_cond_belonging_cluster_thrld}
&contextDistance(c,\ C_i, \ l)\ <\ Th_1 \\
\label{eq_cond_belonging_cluster_min}
&\forall C_j\ \in \ \mathfrak{D}, j \neq i: \ contextDistance(c, C_i, l) \leq  contextDistance(c, C_j, l)
\end{eqnarray}

\noindent where $Th_1$ is defined in section \ref{sec_context_segmentation}. The function $contextDistance(c,\ C_i,\ l)$ is defined in table \ref{tabFunc}. If such a context cluster $C_i$ is found, the satisfactory tracking parameters associated with $C_i$ are considered as good enough for parameterizing the tracking of the current video chunk. Otherwise, the tracking algorithm parameters do not change, the current video chunk is marked to be learned offline later. 

\section{Experimental Results}
\label{secExperimentation}
\subsection{Parameter Setting and Object Detection Algorithm}

The proposed control method has two predefined parameters. The distance threshold $Th_1$ to decide whether two contexts are close enough (sections \ref{sec_context_segmentation} and \ref{sec_para_adaptation}) is set to $0.5$. The minimum number of frames $l$ of a context segment (sections \ref{sec_context_segmentation} and \ref{secContextChange}) is set to $50$ frames. A HOG-based algorithm \cite{corvee10} is used for detecting people in videos.

\subsection{Tracking Evaluation Metrics}

In this experimentation, we select the tracking evaluation metrics used in several publications \cite{kuo}\cite{xing09}\cite{li09}. Let $GT$ be the number of trajectories in the ground-truth of the test video. The first metric \textbf{$MT$} computes the number of trajectories successfully tracked for more than 80\% divided by GT. The second metric \textbf{$PT$} computes the number of trajectories that are tracked between 20\% and 80\% divided by GT. The last metric \textbf{$ML$} is the percentage of the left trajectories.

\subsection{Controlled Tracker}

In this paper, we select an object appearance-based tracker \cite{dpchauIcdp11} to test the proposed approach. This tracker takes as input a video stream and a list of objects detected in a predefined temporal window. The object trajectory computation is based on a weighted combination of five object descriptor similarities: 2D area, 2D shape ratio, RGB color histogram, color covariance and dominant color. For this tracker, the five object descriptor weights $w_k$ ($k$ = 1..5) are selected for testing the proposed control method. These parameters depend on the tracking context and have a significant effect on the tracking quality.

\subsection{Training Phase}
\label{sec_tracker_1_training_phase}

In the training phase, we use 15 video sequences belonging to different contexts (i.e. different levels of density and occlusion of mobile objects as well as of their contrast with regard to the surrounding background, their contrast variance, their 2D area and their 2D area variance). These videos belong to four public datasets (ETISEO, Caviar, Gerhome and PETS) and to the two European projects (Caretaker and Vanaheim). They are recorded in various places: shopping center, buildings, home, subway stations and outdoor.

Each training video is segmented automatically in a set of context segments. In the tracking parameter optimization process, we use an Adaboost algorithm to learn the object descriptor weights for each context segment because each object descriptor similarity can be considered as a weak classifier for linking two objects detected within a temporal window. The Adaboost algorithm has a lower complexity than the other heuristic optimization algorithms (e.g. genetic algorithm, particle swam optimization). Also, this algorithm avoids converging to the local optimal solutions. After segmenting the 15 training videos, we obtain 72 contexts. By applying the clustering process, 29 context clusters are created.

\subsection{Testing Phase}

All the following test videos do not belong to the set of the 15 training videos.

\textbf{1. Caretaker video}

\begin{figure}[t]
\centering
\includegraphics[width=10.5cm]{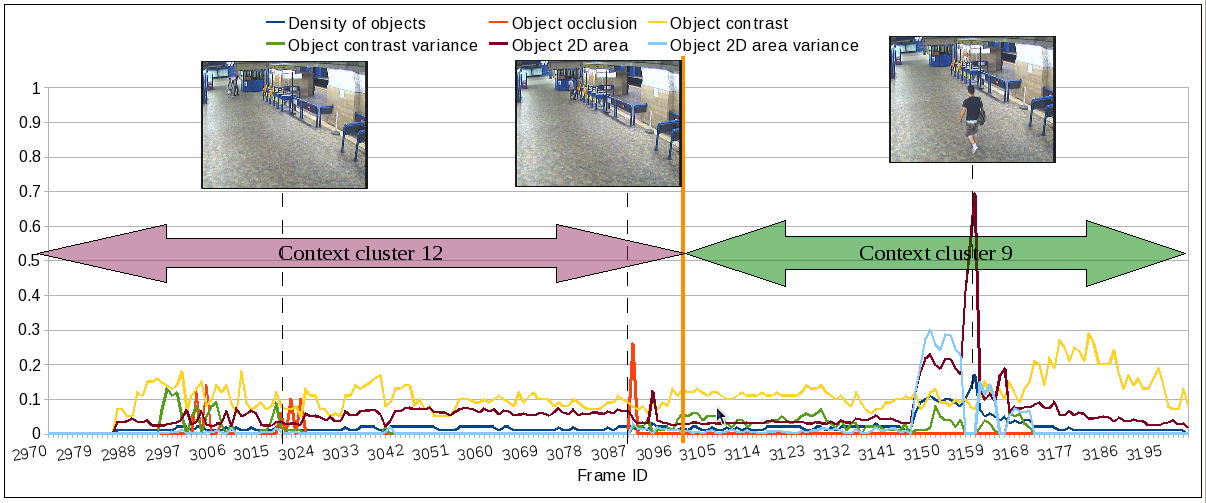}
\caption{Variations of the contextual feature values and of the detected contexts in the subway sequence from frame 2950 to frame 3350.}
\label{fig_caretaker_context_vals_1}
\end{figure}

The first tested video sequence belongs to the Caretaker project\footnote{http://cordis.europa.eu/ist/kct/caretaker\_synopsis.htm} whose video camera is installed in a subway station. The length of this sequence is 1 hour 42 minutes. It contains 178 mobile objects. The graph in figure \ref{fig_caretaker_context_vals_1} presents the variation of contextual feature values and of the detected contexts in the test sequence from frame 2950 to frame 3200. The values of object 2D areas are normalized for displaying. From frame 2950 to 3100, the area and area variance values of objects are very small most of the time (see the brown and light blue curves). The context of this video chunk belongs to cluster 12. In this cluster, the color histogram is selected as the most important object descriptor for tracking mobile objects ($w_3 = 0.86$). This parameter tuning result is reasonable because compared to the other considered object descriptors, the color histogram descriptor is quite reliable for discriminating and tracking objects of low resolution (i.e. low 2D area). From 
frame 3101 to 3200, a larger object appears, the context belongs to cluster 9. For this context cluster, the dominant color descriptor weight is the most important ($w_5$ = 0.52). In this case, the object appearance is well visible. The dominant color descriptor is then reliable for tracking object.

The proposed controller helps to increase the \textit{MT} value from 61.24\% to 71.32\%, and to decrease the value of \textit{ML} from 24.72\% to 20.40\%.

\textbf{2. Caviar Dataset}

\begin{figure*}[t]
\begin{center}$
\begin{array}{cccc}
\includegraphics[width=3cm]{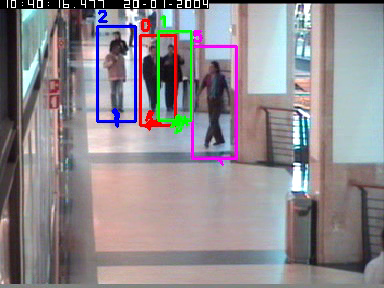} &
\includegraphics[width=3cm]{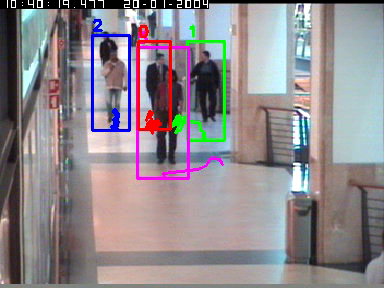} &
\includegraphics[width=3cm]{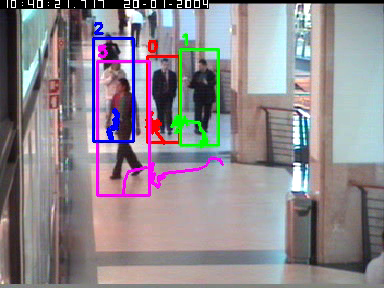} &
\includegraphics[width=3cm]{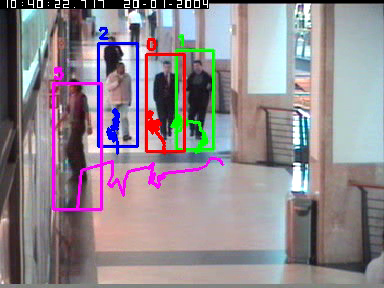} \\
\end{array}$
\end{center}
\caption{\label{fig_caviar_res}Tracking results of four persons in the sequence ShopAssistant2cor (Caviar dataset) are correct, even when occlusions happen.}
\end{figure*}

The Caviar videos are recorded in a shopping center corridor. They have 26 sequences in which 6 sequences belong to our training video set. The other 20 sequences including 143 mobile objects are used for testing. Figure \ref{fig_caviar_res} shows the correct tracking results of four persons while occlusions happen. Table \ref{tab_caviar_result} presents the tracking results of the proposed approach and of some recent trackers from the state of the art. The proposed controller increases significantly the performance of the tracker \cite{dpchauIcdp11}. The $MT$ value increases 78.3\% to 85.5\% and the $ML$ value decreases 5.7\% to 5.3\%. We obtain the best $MT$ value compared to state of the art trackers.

\begin{table}[b]
   \begin{center}
	\begin{tabular}{|l|c|c|c|c|}
		\hline
			  Approaches 	& MT (\%)& PT (\%) & ML (\%)   \\
 		\hline
	Xing et al.\cite{xing09}		& 84.3 	& 12.1 & 3.6	\\
		\hline
	 Li et al.\cite{li09}			& 84.6 	& 14.0 & 1.4  	\\
		\hline
	 Kuo et al.\cite{kuo}			& 84.6 	& 14.7 & \textbf{0.7}	\\
		\hline
Appearance Tracker \cite{dpchauIcdp11} without the proposed controller & 78.3 	& 16.0 & 5.7	\\
		\hline
\textbf{Appearance Tracker \cite{dpchauIcdp11} with the proposed controller} &  \textbf{85.5} & 9.2 &5.3\\
		\hline
	\end{tabular}
\end{center}
\caption{\label{tab_caviar_result}Tracking results for the Caviar dataset. The proposed controller improves significantly the tracking performance. The best values are printed in \textbf{bold}.}
\end{table}

\textbf{3. PETS 2009 Video}

In this test, we use the CLEAR metrics presented in \cite{petsmetric} to compare with other tracking algorithms. The first metric MOTA computes Multiple Object Tracking Accuracy. The second metric MOTP computes Multiple Object Tracking Precision. The higher these metrics, the better the tracking quality is.
We select the sequence S2\_L1, camera view 1, time 12.34 for testing because this sequence is experimented in several state of the art trackers. This sequence has 794 frames, contains 21 mobile objects and several occlusion cases. With the proposed controller, the tracking result increases significantly. Table \ref{tab_pets_result} presents the metric results of the proposed approach and of different trackers from the state of the art. The metric $\overline{M}$ represents the average value of MOTA and MOTP. With the proposed approach, we obtain the best values in all the three metrics.
\begin{table}[t]
   \begin{center}
	\begin{tabular}{|l|c|c|c|c|}
		\hline
			  Approaches 	& MOTA  & MOTP & $\overline{M}$  \\
 		\hline
	 	Berclaz et al. \cite{berclaz06}  	& 0.80	& 0.58 	& 0.69	\\
		\hline
		 Shitrit et al. \cite{shitrit11}  	& 0.81	& 0.58 	& 0.70	\\
		\hline
		 Henriques et al. \cite{henriques11}  	& 0.85	& 0.69 	& 0.77	\\
		\hline
	      Appearance Tracker \cite{dpchauIcdp11} without the proposed controller & 0.62 	&0.63  &0.63	\\
		\hline
\textbf{Appearance Tracker \cite{dpchauIcdp11} with the proposed controller}	& \textbf{0.87} & \textbf{0.72} & \textbf{0.80}  \\
		\hline
	\end{tabular}
\end{center}
\caption{\label{tab_pets_result}Tracking results for the PETS sequence S2.L1, camera view 1, time 12.34. The best values are printed in \textbf{bold}.}
\end{table}

\section{Conclusion and Future Work}

In this paper, we have presented a new control approach for tuning online the tracker parameters. The proposed offline learning phase helps to decrease effectively the computational cost of the online control phase. The experiments show a significant improvement of the tracking performances while using the proposed controller. Although we test with an appearance-based tracker, other tracker categories can still be controlled by adapting the context definition to the principle of these trackers. In future work, we will extend the context notion which should be independent from the object detection quality.

\section*{Acknowledgments}
\noindent This work is supported by The PACA region, The General Council of Alpes Maritimes province, France as well as The Vanaheim, Panorama and Support projects.


\begin{thebibliography}{4}

\bibitem{kuo} Kuo, C.H., Huang, C., Nevatia, R.: Multi-target tracking by online learned discriminative appearance models. In: CVPR (2010)

\bibitem{borji12} Borji, A., Frintrop, S., Sihite, D. N., Itti, L.: Adaptive Object Tracking by Learning Background Context. In: CVPR (2012)

\bibitem{prost} Santner, J., Leistner, C., Saffari, A., Pock, T. Bischof, H.: PROST: Parallel Robust Online Simple Tracking. In: CVPR (2010)

\bibitem{yoon12} Yoon, J.H., Kim, D.Y., Yoon, K.J.: Visual Tracking via Adaptive Tracker Selection with Multiple Features. In: ECCV (2012)

\bibitem{dpchauIcdp11} Chau, D.P., Bremond, F., Thonnat, M.: A multi-feature tracking algorithm enabling adaptation to context variations. In: ICDP (2011)

\bibitem{codebook} Kim, K., Chalidabhongse, T.H, Harwood, D. and Davis, L.: Background modeling and subtraction by codebook construction. In: ICIP (2004)

\bibitem{corvee10} Corvee, E., Bremond, F.: Body parts detection for people tracking using trees of Histogram of Oriented Gradient descriptors. In: AVSS (2010)

\bibitem{petsmetric} Bernardin, K., Stiefelhagen, R.: EvaluatingMultiple Object Tracking Performance: The CLEAR MOTMetrics. In: EURASIP J. on Img and Video Processing (2008)

\bibitem{xing09} Xing, J., Ai, H., Lao, S.: Multi-object tracking through occlusions by local tracklets filtering and global tracklets association with detection responses. In: CVPR (2009)

\bibitem{li09} Li, Y., Huang, C., Nevatia, R.: Learning to Associate: HybridBoosted Multi-Target Tracker for Crowded Scene. In: CVPR (2009)

\bibitem{berclaz06} Berclaz, J., Fleuret, F., Turetken, E., Fua, P.: Multiple object tracking using k-shortest paths optimization. In: TPAMI, vol 33, 1806--1819 (2011)

\bibitem{shitrit11} Shitrit, J., Berclaz, J., Fleuret, F., Fua, P.: Tracking multiple people under global appearance constraints. In: ICCV (2011)

\bibitem{henriques11} Henriques, J.F., Caseiro, R., Batista, J.: Globally optimal solution to multi-object tracking with merged measurements. In: ICCV (2011)


\end{thebibliography}
\end{document}